\RequirePackage{fix-cm}
\documentclass[twocolumn]{svjour3}          
\smartqed  
\usepackage{graphicx}
\usepackage{natbib}
\usepackage{amsmath}
\usepackage[hyphens]{url}
\usepackage[ruled,vlined]{algorithm2e}
\usepackage{multirow}
\usepackage{tabularx}

\usepackage{hyperref}
\usepackage{xcolor}

\journalname{A Novel Data Augmentation Text Generation Approach}
\begin{document}\sloppy

\title{Data Augmentation in Natural Language Processing: A Novel Text Generation Approach for Long and Short Text Classifiers
}

\titlerunning{A Novel Data Augmentation Text Generation Approach} 

\author{Markus Bayer \and Marc-André Kaufhold \and Björn Buchhold \and Marcel Keller \and Jörg Dallmeyer \and Christian Reuter
}


\institute{Markus Bayer \and Marc-André Kaufhold \and Christian Reuter \at
              Technical University of Darmstadt, Darmstadt, Germany \\
              \email{\{bayer, kaufhold, reuter\}@peasec.tu-darmstadt.de}          
           \and
           Björn Buchhold\and Jörg Dallmeyer \at
              CID GmbH, Freigericht, Germany
            \and
            Marcel Keller \at Independent Researcher, Hanau, Germany
}

\date{}

\twocolumn[
    \maketitle
    \begin{@twocolumnfalse}
        \begin{abstract}
            In many cases of machine learning, research suggests that the development of training data might have a higher relevance than the choice and modelling of classifiers themselves. Thus, data augmentation methods have been developed to improve classifiers by artificially created training data. In NLP, there is the challenge of establishing universal rules for text transformations which provide new linguistic patterns. In this paper, we present and evaluate a text generation method suitable to increase the performance of classifiers for long and short texts. We achieved promising improvements when evaluating short as well as long text tasks with the enhancement by our text generation method. Especially with regard to small data analytics, additive accuracy gains of up to 15.53\% and 3.56\% are achieved within a constructed low data regime, compared to the no augmentation baseline and another data augmentation technique. As the current track of these constructed regimes is not universally applicable, we also show major improvements in several real world low data tasks (up to +4.84 F1-score). Since we are evaluating the method from many perspectives (in total 11 datasets), we also observe situations where the method might not be suitable. We discuss implications and patterns for the successful application of our approach on different types of datasets.
        \end{abstract}
    \end{@twocolumnfalse}
]

\section*{Declarations}
\subsection*{\textbf{Funding}}
This work has been co-funded by the German Federal Ministry of Education and Research (BMBF) in the project CYWARN (13N15407) and by the LOEWE initiative (Hesse, Germany) within the emergenCITY center. The calculations for this research were conducted on the Lichtenberg high performance computer of the TU Darmstadt.

\subsection*{\textbf{Conflicts of interest/Competing interests}}
Not applicable.

\subsection*{\textbf{Availability of data and material}}
The availability of data is outlined in appendix \ref{appendix_data_material}.

\subsection*{\textbf{Code availability}}
The code specifics are outlined in appendix \ref{appendix_algorithm}.

\subsection*{\textbf{Ethics}}
Ethics are discussed in appendix \ref{appendix_ethics}.

\section{Introduction} \label{introduction}
Deep learning has attracted considerable attention due to increased computing power in combination with a higher availability of training data for a wide range of problems \citep{Sun2017}. In some learning tasks, especially small data regimes, the development of training data might have a higher relevance than the choice and modelling of classifiers \citep{Banko2001}. To improve classifiers, data augmentation methods have been designed to artificially create training data with specific transformations \citep{Taylor2019}. Current research in data augmentation focuses on deep learning algorithms, which are state of the art for many classification tasks, as they still often suffer from a strong variance regarding the given problem if not enough data is provided. The artificial creation of training data serves as a kind of regularization and thus, simpler solutions are preferred \citep{Zeiler2013, Hu2020}. In addition, imbalance in datasets can be addressed \citep{Zhai_2021,Raghuwanshi2021} and the security of classifiers can be increased by making them resistant to the deception by skillful changes in the input sequences \citep{Miyato2017}. Data augmentation can also help to mitigate the "big data wall" problem, which relates to the fact that smaller companies, research groups and organizations are usually unable to acquire the same volume of data as large corporations \citep{Coulombe2018}. 

Regardless of deep learning, research into artificial data creation can benefit natural language processing (NLP) applications across several domains where training data is scarce or labelling is costly. For example, to enhance the situational awareness of emergency managers, a part of crisis informatics deals with the rapid recognition and subsequent classification of messages and pictures during disasters and emergencies \citep{Alam2020,Kaufhold2020}. Due to the scarcity of financial and personnel resources, emergency services lose valuable time dealing with complicated identification tasks, which eventually can cost lives \citep{Imran2018,reuter_emergency_2016}. This problem of scarcity also applies to small- and medium-sized enterprises (SMEs) when requiring a high quality and volume of labelled data for commercial tasks such as brand analysis or news classification \citep{stieglitzSocialMediaAnalytics2018}. In NLP, there is the difficulty establishing universal rules for transformations of textual data that can be carried out automatically and still maintain the quality of the labelling, which is especially sensitive in domains such as sentiment analysis \citep{MEDHAT20141093}. \citet{Longpre2020} suggest that current pre-training or transfer learning methods in NLP already cover the goals of data augmentation. They argue that augmentation methods that only perturb the input data and do not provide new linguistic patterns fail to increase the classification quality of pre-trained models. 

Thus, we propose a sophisticated generation-based method that overcomes these problems by incorporating new linguistic patterns (i.e., a high grammatical variety) which prove to be useful in combination with pre-trained models. This method does not simply create very similar instances, but very novel ones. Our approach uses two sub methods, whereof one is context-conditional by incorporating parts of the instances (e.g., first words or title) in the generation process and hence suited for long texts, while the other is context independent and suited for short texts. Although there is no clear distinction between long and short texts, we are guided by the 280 character limit (i.e., the length of a message in Twitter), at which most standard NLP data sets would be categorized as small. Thus, we seek to answer three research questions: How can we utilize text generation approaches of data augmentation that achieve a high novelty in the data while preserving the label quality to improve pre-trained machine learning classifiers \textbf{(RQ1)}? In which way is the incorporation of contexts of long text instances in classification problems helpful when using text generation as data augmentation method \textbf{(RQ1.1)}? How is it possible to achieve a quality improvement for classification tasks with short texts when augmenting with text generation \textbf{(RQ1.2)}? 

Contributing to the domain of small data analytics, our results indicate additive accuracy gains of up to 15.53\% and 3.56\% within a constructed low data regime, compared to the no augmentation baseline and another data augmentation technique. As the current track of these constructed regimes is not universally applicable, we also show major improvements in several real world low data tasks (up to +4.84 F1-score). Since we are evaluating the method from many perspectives (in total 11 datasets), we also observe situations where the method might not be suitable. We discuss empirical (i.e., insights into the domain-specific application of small data analytics), practical (i.e., new data augmentation methods based on the GPT-2 language model) and theoretical (i.e., a textual data augmentation basis which is beneficial for pre-trained classification models) implications for the successful application of our approach on different types of datasets.

The paper is structured as follows: After introducing related work on data augmentation, NLP and text generation approaches (section 2), the paper presents both the concept and implementation of a novel text generation data augmentation algorithm (section 3). Furthermore, it presents the method and findings of three rounds of evaluation (section 4) before discussing the implications, limitations and potentials for future research (section 5).

\section{Related Work} \label{related_work}
\subsection{Foundations of Data Augmentation} \label{foundations}

Data augmentation is a machine learning technique that artificially enlarges the amount of training data by means of label preserving transformations \citep{Taylor2019}. First variations of data augmentation can be identified in the well-known LeNet by \citet{LeCun1998}. Using random distortions of training pictures, the MNIST-dataset was ninefold enlarged, so that a better detection of handwritten digits became feasible. 
A relevant term of data augmentation is label preservation, describing transformations of training data that preserve class information \citep{Coulombe2018}. This means that this kind of transformations modifies texts of a given class to other texts that are as well related to this class. In data augmentation research, this is of high relevance because the absence of it would result in the generation of incorrectly classified data. For the most part, an entity replacement within a sentence is sufficient for label preservation in sentiment analysis. However, the random addition of words may result in an alteration of the sentiment. Many researchers loosen the label preservation term. Then, transformations that break the preservation are legitimate as long as the label is adjusted simultaneously. Furthermore, transformations that preserve the right class with a high probability, but not with certainty, may exist. In this understanding, \citet{Shorten2019} designate the probability that the correct label is assigned after a transformation as the safety of a data augmentation method. For example, this uncertainty, if known, could be directly integrated in the label. If unknown, methods like label smoothing can model a general uncertainty.

In NLP, data augmentation is considered a difficult task \citep{Kafle2018} since textual transformations that preserve the label are difficult to define \citep{Kobayashi2018,Wei2019}. Thus many methods have been tried out in research so far. Among them are methods for swapping \citep{Wei2019}, deleting \citep{Huong2020,Qiu2020}, inducing spelling mistakes \citep{Belinkov2018,Coulombe2018}, paraphrasing \citep{Kumar2019}, and replacing of synonyms \citep{Kolomiyets2011,Zhang2015a, xiangLexicalDataAugmentation2021}, close embeddings \citep{Alzantot2018,Wang2015} and words predicted by a language model \citep{Fadaee2017,Jiao2019,Kobayashi2018} on word-level. On a broader level, methods which change the dependency tree \citep{Sahin2019,Xu2016}, perform round-trip-translation \citep{Kruspe2018,Sennrich2016}, or interpolate the input instances \citep{Chawla2002,Zhang2018} are used.
Further studies have dealt with text generation approaches for data augmentation. While \citet{Rizos2019} and \citet{Sun2020} are using recurrent neural networks and generative adversarial networks for short-text augmentation, \citet{Qiu2020} sample instances from a variational autoencoder without length restrictions. Furthermore, \citet{Wang2020} and \citet{Anaby-Tavor2020} used the GPT-2 model for text generation. A more detailed analysis, taxonomy and listing of data augmentation techniques can be found in the data augmentation survey by \citet{bayerSurveyDataAugmentation2021}.

However, challenging many research directions in this area, \citet{Longpre2020} hypothesize that textual data augmentation would only be helpful if the generated data contains new linguistic patterns that are relevant to the task and have not yet been seen in pre-training.

\subsection{Research Gap} \label{research_gap}

Our data augmentation method is inspired by the text generation methods from \citet{Rizos2019}, \citet{Sun2020} and \citet{Qiu2020} while also considering the limitations outlined by \citet{Longpre2020} and seeks to tackle three primary research gaps: 

\begin{enumerate}
    \item Considering short and long texts while maintaining coherence and achieving high novelty;
    \item preserving the labels and quality of the augmentation method;
    \item overcoming the challenge of limited usefulness of textual data augmentation in combination with pre-trained models.
\end{enumerate}

First, in contrast to the works of \citet{Rizos2019} and \citet{Qiu2020}, we consider short as well as long texts as input data instances to our augmentation method, which is covered explicitly by research questions 1.1 and 1.2. Additionally, and in relation to the main research question, our method is characterized by substantial label preservation in combination with the novelty and coherence of the data. At first, the generation is enriched with a special finetuning and prefix addition. Then, a document embedding filter is applied so that instances not associated with the actual class are omitted. Thus, the generation capabilities can be used in full extent while tailoring them to the class data. Furthermore, our experiments are based on the GPT-2-Model by \citet{Radford2018} that achieves very good results in text generation. 

Second, when it comes to usage of the GPT-2 model, \citet{Wang2020} describe no measures for label and quality preservation in their GPT-2 augmentation. \citet{Anaby-Tavor2020} indicate that the GPT-2 model will be further trained and improperly generated instances will be removed. In contrast to our method, the model of \citet{Anaby-Tavor2020} is limited exclusively on sentences as instances and cannot generate coherent text. Furthermore, it uses other safety mechanisms for label preservation. For instance, they use a filter mechanism based on a classifier, that was trained on the class data. This can severely reduce the diversity of the data augmentation method. 

Third, the method proposed in this paper is intended to overcome the issue that textual data augmentation can be of no or small value when used in combination with pre-trained classifiers \citep{Longpre2020}. In contrast to the study by \citet{Longpre2020}, we use the ULMFit model by \citet{Howard2018}. Nevertheless, the model is also pre-trained beforehand and finetuned on each task dataset. As a specialty, we also finetune the encoder with the augmented data for the baseline, making sure that at least the encoder has seen all linguistic patterns before.

\section{Concept and Implementation} \label{concept_implementation}
\subsection{Conceptual Design} \label{design}

The text generation process can be based on any language model with good text generation capabilities. Language models indicate a probability distribution of sequences of words:

\begin{equation}
    P_{\Theta} (w_t | w_{t-k},..,w_{t-1}) ~\forall t
\end{equation}

The model $P_\Theta$  predicts the probability that the current word is $w_t$ given the predecessor words (context) $w_{t-1},… , w_{t-k}$. This enables the  $P_\Theta$  to generate texts. A phrase prefix can be used as context to make the model follow a certain topic by completing exactly this part, abstracting from exact specifics as sampling methods. In addition, a temperature parameter can be introduced to adjust the randomness in the generation of the texts by scaling the logits in the softmax. In order to enable the sensible use of a language model for data augmentation, it has to be ensured that the procedure mainly generates texts which are similar to the training data and, in addition, reflect the respective class (label preservation or safety). In the following, our augmentation method is described, which comprises the specification of three steps for modeling this behavior.

In a first step, the pre-trained model $P_\Theta$  is further trained (finetuning) with the training data $X_c$ of the class $c$ that should be enriched. On the one hand, this enables the model to learn the words, spelling and form of the training data. On the other hand, a bias is generated with regard to the selected class. This means that, for the generation of data, the selected class can be retained more explicitly. In the following, we differentiate between the contextual data augmentation process that is suitable for longer texts and a context independent process for shorter training instances.

In order to further strengthen the safety and label preservation, special “start of text”-tokens are added to each training data in the finetuning input. In the text generation phase, these tokens are used as generation prefixes, signaling the model to generate texts similar to the specific training data. This ensures that the augmented examples are different to each other but remain based on the actual data. If the training data consists of longer texts, i.e. instances containing more than 280 characters, this token can be selected context-based, for example, by appending the first words or the title of each instance (e.g. “$<|startoftext|> (w_1  … w_k )_i$” where $(w_1  … w_k )_i$ is the beginning sequence of the instance $i$). This results in a high diversity of the generated data. Although, if the texts of the dataset are short and no context tokens can be used for appendage, the context independent variant is chosen, where the number of the occurrence of the instance in the training set is concatenated (e.g. “$<|startoftext|> |i|$” where $i$ is the occurrence). As the language model is finetuned on the training data, it can be assumed that it learns to associate the unique token with the respective instance. Thereby, the model is able to recognize the prefix and completes it on the basis of memorization. Ultimately, this implies a strengthened label preservation. However, so that the data is not completely reproduced from memory, uncertainty is introduced in the sampling by adjusting the temperature parameter.

Filtering the generated data is the final heuristic to increase label preservation. For this purpose, document embeddings for each instance of the generated texts and training data of a class are created. The embeddings reflect the content of the respective instances. If in this latent space a data instance from the generated data $X_{gen}$ is too far away from the actual training data $X_c$ of the class to be augmented, it can be assumed that the content differs semantically and/or syntactically, which is why such data is discarded:

\begin{equation}
\begin{aligned}
    X_{filtered} = \{x_i \in X_{gen}  ~| ~dist(Emb(x_i), ~Centroid( \\ Emb(X_c))) < \delta \}
\end{aligned}
\end{equation}

The large generative model is able to interpolate textual content in a sensible and non-trivial way. These capabilities are very promising for data augmentation by creating highly diverse samples that are coherent and contain new linguistic and semantic patterns with regards to the actual data. However, only through the application of the security steps can the model generate class-related data that does not represent the wrong label.

\subsection{Implementation} \label{implementation}
\begin{figure}
  \centering
  \includegraphics[width=\linewidth]{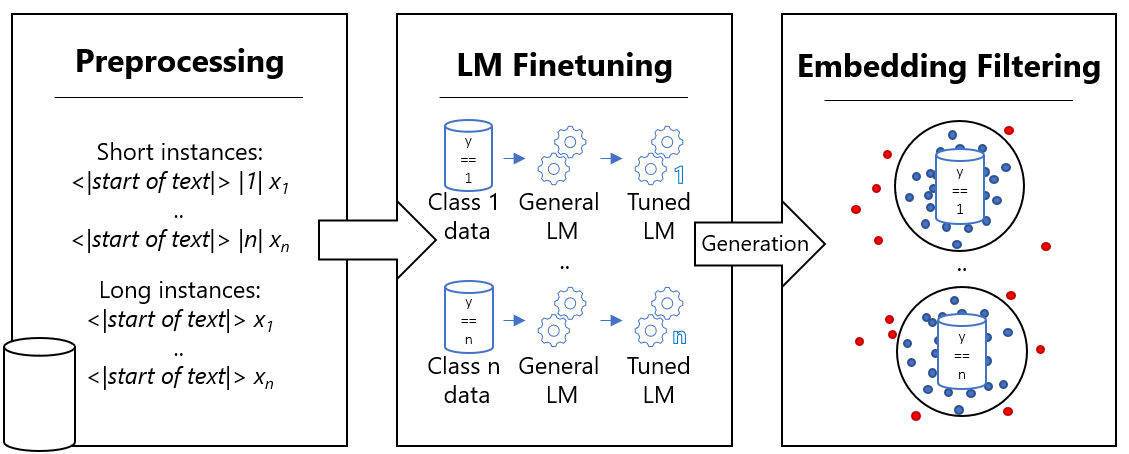}
  \caption[DA Steps]{Three steps to increase the probability of a label preserving instance in GPT-2 text generation (safety increase). In the first step, a contextualized or numbered start token is added to each training instance so that this can be used as a generation prefix for each class after the second step, in which the GPT-2 model is trained further. After the text generation, in a last step a filtration is carried out using BERT document embeddings so that significantly deviant instances are not included.}
  \label{fig:da_steps}
\end{figure}

Figure \ref{fig:da_steps} shows and summarizes the three different steps of the safety enhancement, sorted according to algorithmic order. The class safety of the procedure can be significantly increased by this, although it cannot be completely ruled out that the correct label is obtained. For the implementation, we use GPT-2 by \citet{Radford2018} with 355 million parameters. We used GPT-2 as it is well suited for small data analytics due to its diverse generation capabilities coming from the size. The model is enriched with the three different extensions discussed in the conceptual design section. 

In the first steps, the GPT-2 model is imported and the specific class data is extracted. Subsequently, all instances of this class are given a prefix token (“$<|startoftext|> |\{num\}|$”) and suffix token (“$<|endoftext|>$”). If the training data consists of longer instances with an embeddable context, the “$|\{num\}|$”-field is removed. In all other cases “$\{num\}$” is replaced by the position of the current data instance. Afterwards, the model is finetuned with this data several hundred or thousand epochs, dependent on the dataset size, so that the loss of the model is greatly reduced. This should sufficiently ensure that the model prioritizes the training data in the generation.

Thereafter, texts are generated for each class instance. If the training instances exceed a certain number of words they are considered as long and the token “$<|startoftext|>$” and a specific context of the document (e.g. title or the first words) are added to the generation, else “$<|startoftext|>$” is used in combination with the index of the respective instance. Temperatures between 0.7 and 0.9 should be set in the generation step \citep{Woolf2019}, whereby a higher number represents greater randomness/creativity. In the last step of the procedure, the generated data is filtered. This is done by using Sentence-BERT \citep{Reimers2019} to create document embeddings of the data. Generated instances that are, according to a manually set threshold, too far away from the centroid of the correct data, are deleted from the result set. To minimize this interaction, a predefined value (e.g. 0.3) is set and the algorithm displays the 10 furthest instances that are still within this threshold. Depending on how many instances are wrong, the threshold is moved further and the process is started again. For example, if there is one false instance in the set of 10 generated instances, the threshold is increased slightly. If there are only true instances, the threshold is decreased. This is done until a meaningful parameter is found. 

The algorithms for long and short instances are given in appendix \ref{appendix_algorithm}.
%
\section{Evaluation} \label{evaluation}

\subsection{Selection of Application Domains}
To conduct our evaluation, we selected the three cases of sentitment analysis, news classification and crisis informatics. First, sentiment analysis is, according to \citet{MEDHAT20141093}, the analysis of opinions, attitudes and emotions toward individuals, events or topics. It is a very common NLP task and used in a broad variety of applications, as, for example, the decision making process of organisations and individuals is increasingly dependent on public opinions \citep{Liu2012}. As it is part of several textual machine learning benchmarks and highly used in research, it is a sensible task for the experiments of this paper.
To place this experiment in the setting of small data analytics, an artificially downsampled version will be used.

Second, due to the constantly growing number of news items and their information sources, it is becoming more and more complicated keeping track of topics and finding specific articles \citep{Carreira2004}. Classifiers are used to automatically divide news into predefined classes \citep{Krishnalal2010}. However, news are highly dynamic, so that the source domains are constantly shifting and new classes emerge. As an extreme example in this context, one can compare the news landscape before and after COVID-19 occurred. Such shifts and newly emerging topics have as result that new data has to be labeled all the time, leading to classifiers with small data bases. While this is already sufficient to form a focus in this study, we are also interested in exploring classification as well as data augmentation for large texts, as this receives little attention in research.

Third, the research field of crisis informatics draws on computing and social science perspectives to study the ways in which ICT enables, constrains, and mediates human practices related to crisis and disaster \citep{Soden2018}. Besides the topics of crisis communication, community interaction, and inter-organizational collaboration \citep{reuter_crisis_2012}, crisis informatics examines the application of machine learning to reduce the information overload of irrelevant information, extract useful information from social media (e.g., eyewitness reports, multimedia files), and enhance information quality for both an improved situational awareness and decision making of emergency services \citep{kaufhold_information_2021}. Despite the considerable volumes of social big data disseminated during large-scale emergencies, there is a class imbalance since only a small number of social media posts contribute to situational awareness \citep{Alam2020}. Furthermore, emergency services such as fire or police departments often lack the financial and personnel resources to engage in comprehensive dataset labelling tasks \citep{Imran2018,reuter_emergency_2016}. In contrast, there might be a lack of available raw data in small-scale and uncommon types of emergencies, qualifying crisis informatics as an interesting application field for data augmentation and small data analytics.

\subsection{Model and Datasets} \label{model_datasets}
In accordance with the research questions, the previously conceptualized and implemented data augmentation methods are evaluated in this chapter based on a constructed low data regime with the SST-2 dataset (Results I) and real world low data regimes regarding topic classification of long (Results II) and short texts (Results III). We used the ULMFit model by \citet{Howard2018} that consists of a pre-trained encoder coupled with a linear pooling network and a softmax output. The encoder is finetuned for each task on all the available task specific data (including the augmented instances). Then, the whole network is trained on a supervised task. 

For the evaluation of the context independent method with short texts, we focused on sentiment analysis and the classification of crisis Twitter data. Sentiment analysis will be performed with subsampled SST-2 \citep{Socher2013} datasets to simulate a low data regime on a standardized dataset, similar to \citet{Hu2019} and \citet{Kumar2020}. As these constructed conditions are restricted in their real world applicability, we perform further evaluations with real world low data regimes. 

The crisis classification tasks have very limited resources, as described by \citet{Kaufhold2020}. The first three datasets from \citet{Olteanu2015} are labeled according to whether or not they are informative on the specific topics that are related to the Boston Bombings, the Bohol Earthquake and the West Texas Explosions in 2013. The other two datasets from \citet{Schulz2017} consist of city-specific Twitter posts that are labeled as incident-related or not. 

For the evaluation of the contextual method with long texts, we gathered news articles for topic classification from 2019 and 2020. For the contextual start token we used the respective titles. Topic classification in the news context also faces the problem of few data instances, because the news and dependent topics are often very dynamic and research data is limited. In our case, for every topic, expert groups of two people decided whether the inspected article is relevant to the topic. Furthermore, there is a labeling guideline for every topic so that disagreements are excluded if possible (see appendix \ref{appendix_dataset}). The topics include three economic issues: layoff, management change (MC) and mergers and acquisitions (M\&A) as well as two crisis issues: flood and wildfire. 

Before augmenting the data, a fifth of every set was split into a holdout set. 


\subsection{Evaluation Settings and Pre-Evaluation of Hyperparameters} \label{hyperparameters}
All data augmentation methods are compared against a baseline with the help of 10-fold training executions. Additionally, the sentiment analysis results are compared with the EDA data augmentation method from \citet{Wei2019}. For the sentiment analysis task we augment both classes, while for the others we augment the minority class. 

The text generation process offers various possibilities for hyperparameter optimization, which we evaluated with different datasets to avoid overfitting. Ablation studies are shown on the sentiment dataset (section \ref{ablation_studies}). 
During generation, the model has a temperature parameter which, the larger it is chosen, the more creative the texts will be and new linguistic patterns occur. However, a too high value can lead to instances that are not topic related. A value that is too low can mean that the model repeats itself very often, while a value too high can result in a loss of the actual theme in the texts \citep{Khan2019}. 
According to the author of the implementation used in this paper, the most suitable value is between 0.7 and 0.9 \citep{Woolf2019}.
In a evaluation with the management change topic, 0.7 was most suitable for the existing case of application, compared to 0.8 and 0.9.

The filtering of generated documents is of importance in this process, as GPT-2 can generate novel instances that may have no relation to the actual class. The parameter of this filtering was chosen individually so that the ten most distant documents would still be marked accordingly to the class. In the ablation studies of section \ref{ablation_studies} we show that this filtering is necessary to achieve the high results of this method.

Another aspect that can be added to hyperparameter optimization concerns the number of documents to be generated per training instance. As this is a very important factor we consider it in the evaluations of section \ref{ablation_studies}.
More details on the hyperparameters of the model can be found in the appendices.

\subsection{Results I: Sentiment Analysis (context independent method)} \label{results_1}

\begin{table*}[t]
\caption{Accuracy of the non-contextual text generation process, EDA \citep{Wei2019} and the baseline with regard to different SST-2 subsamples (10 runs).}
\label{table:sentiment_eval}
\centering
\begin{tabular}{l@{\hspace{1em}}l@{\hspace{1em}}l@{\hspace{1em}}l@{\hspace{1em}}l@{\hspace{1em}}}
\hline
Dataset   & Run      & Baseline & EDA    & Text Gen \\ \hline
SST-2 100 & AVG (SD) & 0.5581 (0.0463)   & 0.6934 (0.0124) & \textbf{0.7134} (0.0207)         \\
          & Best     & 0.6226   & 0.7139 & \textbf{0.7495}           \\ \hline
SST-2 300 & AVG (SD) & 0.7241 (0.0119)   & 0.7217 (0.0047) & \textbf{0.7402} (0.0067)         \\
          & Best     & 0.7417    & 0.7295 & \textbf{0.7534}          \\ \hline
SST-2 500 & AVG (SD) & 0.7505 (0.0077)   & 0.7534 (0.0074) & \textbf{0.7598} (0.0126)          \\
          & Best     & 0.7651   & 0.7671 & \textbf{0.7754}          \\ \hline
SST-2 700 & AVG (SD) & \textbf{0.7646} (0.0054)  & 0.7578 (0.0038) & 0.7627 (0.0066)          \\
          & Best     & 0.7705   & 0.7632 & \textbf{0.7754}         \\ \hline
\end{tabular}
\end{table*}



The results considered good from a human point of view are also reflected in the quantitative evaluation results, that are presented in Table \ref{table:sentiment_eval}. The proposed data augmentation method has almost in every case better results than the baseline and the EDA method by \citet{Wei2019}. Particularly it gains the best improvements the less data is available (additive up to 15.53\% and 3.56\% compared to the baseline and EDA). However, also  with the most data the augmentation method reaches in the best run additive performance improvements of 0.49\% and 1.22\% compared to the baseline and the EDA method.

The method has the highest improvements if less data is available because the prior knowledge of the GPT-2 model is most effective there. The model also produces well written instances which is not the case with the EDA method that sometimes fails to improve the baseline. Furthermore, the rationale of low performing data augmentation methods of \citet{Longpre2020} comes into play. In contrary to the EDA method, the proposed augmentation algorithm enriches the training data with new linguistic patterns that have not already been seen by the encoder. 

\subsubsection{Ablation Studies}\label{ablation_studies}

Further, we want to show an excerpt of the relevant results from our ablation evaluations (Figure \ref{fig:ablation_studies}). First, we tested different augmentation sizes that are shown on the SST-2 100 dataset. We limited this evaluation to a maximum of 10 augmentation samples per instance, as the higher numbers demand more computing time. It is evident that the higher the size, the better the results are. 
A human inspection indicates that even higher numbers might not be as beneficial since the repetition within the samples per instance increases. The human inspection process is detailed in Section \ref{appendix_data} in the Appendix.

Furthermore, we also removed the steps of the augmentation process to see the contributions of each. In a first testing case, we did not include the number of the instance in the finetuning and generation phase (indicated by "w/o n." in Figure \ref{fig:ablation_studies}). The decrease in the average accuracy by 5.42 points shows that this component is highly important for the whole augmentation process. This also applies to the last step of the augmentation method (indicated by "w/o f." in Figure \ref{fig:ablation_studies}). Without the manual filtering the average accuracy is reduced by 2.64 points. This insight was already noticed when the filtering parameter was chosen in every task and some instances seemed to be unrelated to the class. 

In summary, this indicates that all the steps of the augmentation process need to be included to reach the highest scores. The next evaluation studies are based on this best combination.

\begin{figure}
  \centering
  \includegraphics[width=\linewidth]{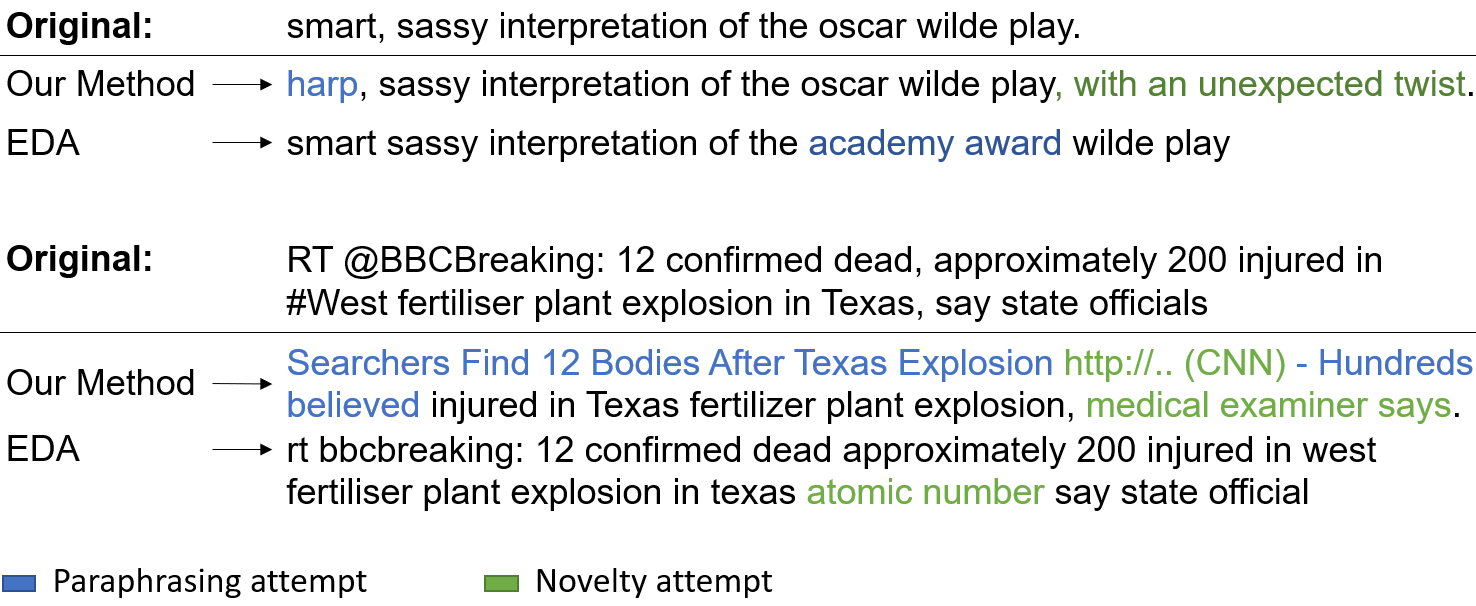}
  \caption[First error analysis example]{Two instances and their transformations by EDA and our method. The first is from the SST-2 task \ref{results_1} and the second from the West Texas Explosion task \ref{results_3}. Text passages where each augmentation method attempted to paraphrase the original instance are highlighted in blue; attempted interpolations or introduction of novelties are highlighted in green.}
  \label{fig:error_analysis}
\end{figure}

Furthermore, we conducted an error analysis by inspecting the generated instances of our technique and comparing them to the EDA method. For our method, it is not clear from which original instance the generation originates, as it could be an interpolation of more than one instance. Nevertheless, we tried to find the closest original instances by measuring the resemblance by Levenshtein distance.

We selected some insightful examples, which are presented in Figure \ref{fig:error_analysis}. The first example shows that EDA is able to keep the label, but substitutes a word that should not be substituted (\textit{oscar wilde play $\rightarrow$ academy award wilde play}). On the other hand, our method is able to find suitable words for the replacement and expands the original instance in a meaningful way without distorting the label or the general content. Nevertheless, while we did not see this case in our analysis, it might be possible that the language model is not able to infer the label during finetuning and that the augmentation would change the label. A similar case can be seen regarding the Dublin task of Section \ref{results_3}, where we assume that the model was not able to infer the label due to a very high diversity in the instances and the label space. While the model does not incorrectly change the label, we see that more than 50\% of the generated instances are ``????????????''. The more data is generated, the more such instances are created and the content-rich instances are repeated.

The second example displayed in Figure \ref{fig:error_analysis} shows that our method brings a high variation with sensible content but also reformulates ``say state officials'' to ``medical examiner says'' which might not be correct. For the task at hand, it is not decisive but it might be with regard to very specific tasks. Moreover, EDA sometimes changes the instance so that it is grammatically incorrect, as it can also be seen in this example. This can be a problem, for example, when using language models that do not expect a specific expression, such as ``atomic number'', in the context of this instance. Furthermore and even worse, it might happen that the method deletes an essential word, like ``not'', in the sentence ``This movie was not bad'', creating an instance with a wrong label when used for sentiment classification.

These examples show that the method proposed in this paper is able to create high-quality and diverse instances. The EDA method instead sometimes creates instances that are wrong or not fitting into the context due to its random character, which is especially critical when using pre-trained language models. Section \ref{appendix_data} in the Appendix provides further examples and analyses of our data augmentation method. It is shown, for example, that the model sometimes completely replicates instances of the training data. This property is not necessarily bad, as it can at least be seen as a sophisticated oversampling method that clones the very important data.

\begin{figure}
  \centering
  \includegraphics[width=\linewidth]{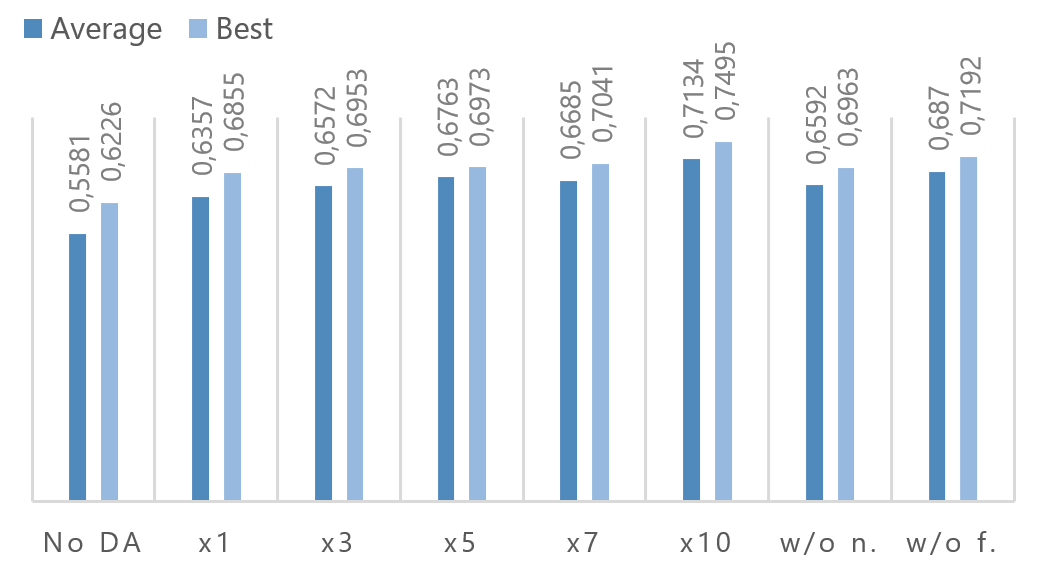}
  \caption[Ablation]{Evaluation of different augmentation sizes and the omission of the numbering token (w/o n.) and the filtering (w/o f.) step on the SST-2 100 dataset (10 runs).}
  \label{fig:ablation_studies}
\end{figure}

\subsection{Results II: News Classification (context dependent method)} \label{results_2}
A qualitative inspection of the data generated for the news dataset shows that high-quality, coherent, and diverse texts have been generated for the different article titles. Nearly all instances had a clear reference to the actual class. In the filtering step, primarily instances where the GPT-2 model frequently repeated words were sorted out.

\begin{table}
\caption{Accuracy and F1 scores of the contextual text generation process and the baseline with regard to the five news article topics (10 runs).}
\label{table:long_text_eval}
\centering
\begin{tabular}{l@{\hspace{1em}}l@{\hspace{1em}}l@{\hspace{1em}}l@{\hspace{1em}}}
\hline
Dataset                            & Run    & Accuracy       & F1               \\ \hline
Layoff                             & AVG (SD) & 0.8350 (0.015) & 0.7695 (0.012)   \\
                                   & Best   & 0.8545         & 0.7905           \\
\textbf{with DA}                            & AVG (SD) & \textbf{0.8706} (0.009) & \textbf{0.8179} (0.011)   \\
                                   & Best   & \textbf{0.8848}         & \textbf{0.8354}           \\ \hline
MC                                 & AVG (SD) & 0.8760 (0.005) & 0.7217 (0.021)   \\
                                   & Best   & 0.8809         & 0.7627           \\
\textbf{with DA}                            & AVG (SD) & \textbf{0.8853} (0.015) & \textbf{0.7559} (0.031)   \\
                                   & Best   & \textbf{0.9077}         & \textbf{0.8052}           \\ \hline
M\&A                               & AVG (SD) & 0.8926 (0.005) & 0.6953 (0.011)   \\
                                   & Best   & \textbf{0.8999}         & 0.7075           \\
\textbf{with DA}                           & AVG (SD) & \textbf{0.8975} (0.003) & \textbf{0.7095}   (0.011) \\
                                   & Best   & \textbf{0.8999}         & \textbf{0.7266}           \\ \hline
Flood                             & AVG (SD) & \textbf{0.8462} (0.007) & 0.8779 (0.007)   \\
                                   & Best   & 0.8540         & 0.8867           \\
\textbf{with DA}                            & AVG (SD) & 0.8408 (0.010) & \textbf{0.8804} (0.006)   \\
                                   & Best   & \textbf{0.8594}         & \textbf{0.8931}           \\ \hline
Wildfire                           & AVG (SD) & 0.9287 (0.016) & 0.9253 (0.017)   \\
                                   & Best   & \textbf{0.9419}         & \textbf{0.9395}           \\
With DA                            & AVG (SD) & \textbf{0.9312} (0.007) & \textbf{0.9297}   (0.006) \\
                                   & Best   & 0.9395         & 0.9375           \\ \hline
\end{tabular}
\end{table}

The classification results of our contextual approach are presented in Table \ref{table:long_text_eval} in comparison to the baseline. Looking at the overall evaluation across all topics, it can be seen that the classifiers also achieve very good results through the contextual data augmentation method. Especially for the economic tasks, relative increases in the maximum F1 value of over 4\% for the MC and layoff topic and 2\% for the M\&A task can be recorded. The M\&A evaluation illustrates the importance of the F1 measure. While good improvements in accuracy were also achieved for management change and layoff, there is virtually no change in the M\&A topic. However, the more significant F1 measure increases substantially. A small improvement is also evident for the flood topic (+0.64 with the maximum F1 value). For the wildfire topic, the text generation approach does not achieve better results. The values are almost the same as for the baseline. However, the standard deviations of the results from the baseline are more than twice as high as with the data augmentation method.

It can be ascertained that the contextual text generation approach is very well suited for the present topic classification tasks. A possible bias regarding the economic topics could be attributed to the pre-trained GPT-2 model. The model was trained with documents from outgoing Reddit links. There may be comparatively few crisis data among the approximately 8 million documents so that the model is less able to represent this topic area. Another possible explanation for the smaller values of the crisis topics is the already very high classification quality of the two tasks. The flood topic, which was chosen because of the poorer classification quality compared to the wildfire topic, still exceeds the three economic topics in the F1-measure. The wildfire topic is by far the best task for the classifiers (about +5\% F1-measure compared to the flood task). An improvement of the two topics may not be possible anymore because the dataset can contain errors as some labelling decisions are difficult and subjective. 

\subsection{Results III: Crisis Informatics (context independent method)} \label{results_3}
In the last section, we stated that GPT-2 might be less usable for crisis data. Since the usage of machine learning in crisis situations is very promising and getting good models is an ongoing issue due to little data and the challenge of domain adaption, we further examine this consideration by focusing only on crisis data for the second evaluation. 

\begin{table}
\caption{Accuracy and F1 scores of the non-contextual text generation process and the baseline with regard to the five crisis Twitter topics (10 runs).}
\label{table:short_text_eval}
\centering
\begin{tabular}{l@{\hspace{1em}}l@{\hspace{1em}}l@{\hspace{1em}}l@{\hspace{1em}}}
\hline
Dataset              & Run    & Accuracy       & F1               \\ \hline
Boston               & AVG (SD) & 0.7886 (0.019) & 0.7344 (0.030)   \\
Bombings             & Best   & 0.8062         & 0.7720           \\
\textbf{with DA}              & AVG (SD) & \textbf{0.8003} (0.021) & \textbf{0.7588} (0.024)   \\
                     & Best   & \textbf{0.8311}         & \textbf{0.7979}           \\ \hline
Bohol                & AVG (SD) & 0.9097 (0.014) & 0.8857 (0.021)   \\
Earthquake           & Best   & 0.9302         & 0.9126           \\
\textbf{with DA}              & AVG (SD) & \textbf{0.9238} (0.011) & \textbf{0.9062} (0.015)   \\
                     & Best   & \textbf{0.9399}         & \textbf{0.9277}           \\ \hline
West Texas           & AVG (SD) & 0.8486 (0.020) & 0.8340   (0.025) \\
Explosion            & Best   & 0.8804         & 0.8765           \\
\textbf{with DA}              & AVG (SD) & \textbf{0.8755} (0.010) & \textbf{0.8721} (0.011)   \\
                     & Best   & \textbf{0.9004}         & \textbf{0.8970}           \\ \hline
\textbf{Dublin}               & AVG (SD) & \textbf{0.9893} (0.002) & \textbf{0.9199} (0.015)   \\
                     & Best   & \textbf{0.9912}         & \textbf{0.9351}           \\
with DA              & AVG (SD) & 0.9858 (0.002) & 0.8945   (0.014) \\
                     & Best   & 0.9878         & 0.9116           \\ \hline
New York             & AVG (SD) & 0.9302 (0.016) & 0.8428   (0.027) \\
City                 & Best   & \textbf{0.9463}         & \textbf{0.8701}           \\
with DA              & AVG (SD) & \textbf{0.9346} (0.003) & \textbf{0.8472} (0.007)   \\ 
                     & Best   & 0.9385         & 0.8555          \\ \hline
\end{tabular}
\end{table}

Inspecting the newly generated data, we see that the model often produces identical outputs for different runs. The generated data of the context dependent models of the second evaluation is clearly more diverse. However, the context independent method also seems to have the potential to perform very well, as Table \ref{table:short_text_eval} shows. It is especially beneficial for the classification tasks of Olteanu et al. (2015) (first three datasets). The averages and best runs outperform the baseline additively by 2.1\% to 3.8\% and 1.5\% to 2.5\% in the F1-measure. For the tasks of \citet{Schulz2017}, however, the method does not provide any substantial improvements regarding the direct scores. For the Dublin dataset, the augmentation method even seems to have a negative effect. Yet, a decrease in the F1 standard deviation is achieved for every task.

Although we can confirm the assumption of the sentiment analysis that the context independent variant of the augmentation method creates less diverse instances, it analogously has a very positive impact on the classification quality when applied to real world low data regimes. Especially the narrowly defined problems by \citet{Olteanu2015} are well suited, leading to the consideration that the difference of the results lies in the nature of the problems. While the first three tasks are bond to a special crisis event, the two other tasks are just incident related with no other focus than the respective city. It may be that these two tasks are too broadly defined with very different instances so that the model was not able to properly finetune to generate sensible instances. Apart from that, the gain in robustness is clearly visible because a decrease in the F1 standard deviation is achieved on each task. The evaluation also shows that our data augmentation methods can achieve good results not only on economic topics.
\section{Discussion and Conclusion} \label{discussion}

While there are numerous beneficial data augmentation methods in computer vision, textual transformations are more difficult to define \citep{Kobayashi2018,Wei2019} and often result in mixed results \citep{Longpre2020}. In order to address these issues, we presented two data augmentation methods for long and short texts based on text generation techniques to enhance the knowledge base on small data analytics. Our results on the 11 datasets, which are listed in an aggregated form in Table \ref{table:comprehensive_eval}, contribute to answering the following research questions.

\begin{table}
\caption{Average and maximum F1 performance deltas of the data augmentation models in comparison to their respective baseline counterparts across all datasets.}
\label{table:comprehensive_eval}
\centering
\begin{tabular}{lll}
\hline
Dataset                & Delta   Avg.  & Delta Max   \\ \hline
SST-2 (100) - Acc.     & +15.53\%        &  +12.69\%      \\
SST-2 (300) - Acc.     & +1.61\%         &  +1.17\%       \\
SST-2 (500) - Acc.     & +0.93\%         &  +1.03\%       \\
SST-2 (700) - Acc.     & -0.19\%         &  +0.49\%       \\ \hline
Layoff - F1            & +4.84\%         & +4.49\%        \\
MC - F1                & +3.42\%         & +4.25\%        \\
M\&A - F1              & +1.42\%         & +1.91\%        \\
Flood - F1             & +0.25\%         & +0.64\%        \\
Wildfire - F1          & +0.44\%         & –0.20\%        \\ \hline
Boston   Bombings - F1 & +2.44\%         & +2.59\%        \\
Bohol   Earthquake - F1 & +2.05\%        & +1.51\%        \\
West   Texas Explosion - F1 & +3.81\%    & +2.05\%        \\
Dublin - F1            & –2.54\%         & –2.35\%        \\
New York   City - F1   & +0.44\%         & –1.46\%        \\ \hline
\end{tabular}
\end{table}

\textbf{How can we utilize text generation approaches of data augmentation that achieve a high novelty in the data while preserving the label quality to improve pre-trained machine learning classifiers (RQ1)?} 
We proposed two data augmentation methods that are based on text generating language models. We constructed three different steps (see Figure 1) for ensuring a high label preservation within the transformations of these models. As a first step, the data was primed by a special token. This token can signal the model to generate training data for this class in the generation phase. In order to ensure that the model is familiar with the class data and the token, finetuning was carried out with the prepared data in the second step. After the generation, filtering based on the BERT document embeddings \citep{Reimers2019} formed the final step. In our experimental setup, we used a pre-trained encoder and finetuned it on the various tasks, including the augmented data, to face the challenge of creating a sophisticated data augmentation method that also performs well on pre-trained models. The two derived methods achieved very good results in the evaluation phase with several performance gains and reductions of the standard deviations.

When evaluating the utility of the algorithm, however, further criteria must be considered. While it is rather easy to embed the augmentation method into a classification process, the GPT-2 model needs some time to be executed. The generation of one example of a long dataset took about 10-30 seconds. Noise-inducing methods such as EDA take much less than a second to complete an instance, as they only perform simple operations such as changing the order of words, deleting some words, or generating misspellings \citep{Huong2020,Qiu2020, Belinkov2018,Coulombe2018}. However, text generation methods such as ours are limited by the time required by the generation process. Nevertheless, compared to the time it takes a human to label a new instance, our method is still very advantageous. The time required can also be significantly reduced, for example, by using a different language model that is faster. Furthermore, the used GPT-2 model is mainly limited to English, making it less usable for multilingual tasks. However, this can also be mitigated by using another language model as the proposed method is suitable for different language models.

\textbf{In which way is the incorporation of contexts of long text instances in classification problems helpful when using text generation as data augmentation method (RQ1.1)?} 
When dealing with long texts, we decided to integrate a context-based token in the generation phase so that the generated texts are more explicit for the respective instance and highly diverse among all instances. A closer look into the generated samples from the evaluation phase confirmed this assumption. Furthermore, the generated instances seem to be very coherent and related to the task at hand, due to the strengths of the GPT-2 language model. More importantly, four of the five tasks could be improved by including the newly generated and filtered instances. This led to an additive increase in the average and maximum F1 value of up to 4.8\% and 4.5\% respectively. However, we noticed that the augmentation technique could not improve the classification results when the classifier already performs very well without additional data.

\textbf{How is it possible to achieve a quality improvement for classification tasks with short texts when augmenting with text generation (RQ1.2)?} 
For classification tasks with short texts, a context-based token integration is not possible, wherefore we included the number at which the respective instance occurred in the finetuning. On closer inspection of the newly created instances, several duplicates were found. This did not have negative implications for the evaluation, since repeating some training examples resembles the process of simple oversampling. It may even be interpreted as a more sophisticated version, where some examples are completely new and the others are oversampled from the most fitting data points. Accordingly, a great performance gain could be achieved in the constructed and real world low data regimes, leading to improvements of up to 15.53 and 3.81 points respectively. We noticed that this augmentation method was not suitable for two special real world tasks. We hypothesize that these two tasks are too broadly defined on the rationale that the GPT-2 model is not able to infer the right context just based on the finetuning of the data.

\subsection{Empirical, Practical and Theoretical Contributions} \label{contributions}

Considering our findings, the study revealed practical, theoretical and empirical contributions:

\textbf{New data augmentation methods based on the GPT-2 language model.} The evaluation results of the data augmentation methods indicated that the GPT-2-model, in combination with three safety steps, can achieve a considerable improvement in the text classification tasks (see Table \ref{table:sentiment_eval} to \ref{table:short_text_eval}). In contrast to the similar approaches of \citet{Wang2020} and \citet{Anaby-Tavor2020} that utilized GPT-2 for text generation too, our method is more generally applicable and offers more safety steps. \citet{Wang2020} describe no measures for label preservation in their approach, and \citet{Anaby-Tavor2020} only enable data augmentation for instances consisting of one sentence. Furthermore, we include a filtering mechanism which includes the human expertise, strongly increasing the diversity without much supervision.  The advantages of the text generation approach proposed here facilitate a wide use, qualifying it as basic element for further adaption in prospective classification applications.

\textbf{A textual data augmentation basis which is beneficial for pre-trained classification models.} \citet{Longpre2020} show that data augmentation might not be helpful when dealing with state-of-the-art pre-trained models. This seems logical from a theoretical perspective since pre-training and the transfer to new tasks also follow the goal of reducing the amount of necessary training data. In order to get an enrichment anyway, sophisticated augmentation methods are needed, which should provide unseen linguistic patterns that are relevant to the task \citep{Longpre2020}. We addressed this issue by leveraging the GPT-2 model that was trained with more than 8 million web pages. This gives the great potential to include new linguistic patterns in the generated data (example instances can be found in appendix \ref{appendix_data}). For creating task relevant data, we derived three steps that increase the possibility of class related content. In the evaluation we showed that the proposed method is able to improve the pre-trained encoder model. In contrast to \citet{Longpre2020}, we did not test the method on a transformer model. However, we trained the pre-trained ULMFit encoder for both testing cases (no augmentation and augmentation) with the augmented data so that for the encoder no data is unseen beforehand.

\textbf{Empirical insights into the domain-specific application of small data analytics.} In this work, we gathered new empirical insights into the application of data augmentation in the research domains of sentiment analysis, news classification, and crisis informatics. In crisis informatics, various studies have examined the use of domain adaptation, transfer learning, active learning, and online learning to reduce the labeling effort \citep{Imran2018,Kaufhold2020,Nguyen_2017}. However, few research has examined the application of textual data augmentation for crisis management \citep{Wang2020}, which we enhance by the evaluation and interpretation of seven augmented datasets.
For sentiment analysis we constructed a low data regime, like \citet{Kumar2019} and \citet{Hu2019}. Small data analysis research is gaining popularity and there is a need to establish datasets that can be used, understood and compared by all types of researchers. We strengthen this research direction by basing our evaluation on this dataset. Furthermore, with the five news classification datasets we are exploring the important topic of long text classification, that is often neglected in research. News classification is mostly done with short descriptions of the articles only, as in the AG News dataset \citep{Zhang2015a}. 

\subsection{Limitations and Outlook} \label{limitations}


During the evaluation, a restriction was made with regard to the language model used. While applicable with other language models, it is not clear if the performance gain remains the same. The GPT-3 model by \citet{Brown2020} seems to be the next sensible choice for increasing the results. However, the finetuning step that is likely to be necessary is currently not possible due to the high resource utilization of the model. Nonetheless, because of its size and linguistic expressiveness, it may be especially helpful to address the challenge stated by \citet{Longpre2020} in which pre-trained models might not gain any improvement from data augmentation. In relation to the study by \citet{Longpre2020}, future research could test the proposed method with transformer models. It would also be interesting to see how smaller language models perform, that may be much faster. In addition, there might also be an option to fully automate the filtering step, which further increases the universal usability, even if the human effort is already very low now.

Despite many efforts in data augmentation, the big data wall problem addressed at the beginning is still of great relevance. However, if in the future, according to various assumptions, very large models, such as GPT-3 by \citet{Brown2020}, are better able to solve these problems, the high resource wall problem opens up, which only allows large companies to train and use these models. 

\begin{acknowledgements}
This work has been co-funded by the German Federal Ministry of Education and Research (BMBF) in the project CYWARN (13N15407) and by the LOEWE initiative (Hesse, Germany) within the emergenCITY center. The calculations for this research were conducted on the Lichtenberg high performance computer of the TU Darmstadt.

\end{acknowledgements}

%
%

\bibliographystyle{spbasic}      
\bibliography{references}   

\begin{thebibliography}{66}
\providecommand{\natexlab}[1]{#1}
\providecommand{\url}[1]{{#1}}
\providecommand{\urlprefix}{URL }
\expandafter\ifx\csname urlstyle\endcsname\relax
  \providecommand{\doi}[1]{DOI~\discretionary{}{}{}#1}\else
  \providecommand{\doi}{DOI~\discretionary{}{}{}\begingroup
  \urlstyle{rm}\Url}\fi
\providecommand{\eprint}[2][]{\url{#2}}

\bibitem[{Şahin and Steedman(2018)}]{Sahin2019}
Şahin GG, Steedman M (2018) {Data Augmentation via Dependency Tree Morphing
  for Low-Resource Languages}. In: Proceedings of the 2018 Conference on
  Empirical Methods in Natural Language Processing, \doi{10.18653/v1/d18-1545}

\bibitem[{Alam et~al.(2020)Alam, Ofli, and Imran}]{Alam2020}
Alam F, Ofli F, Imran M (2020) Descriptive and visual summaries of disaster
  events using artificial intelligence techniques: case studies of hurricanes
  harvey, irma, and maria. Behaviour \& Information Technology 39(3):288--318,
  \doi{10.1080/0144929X.2019.1610908}

\bibitem[{Alzantot et~al.(2018)Alzantot, Sharma, Elgohary, Ho, Srivastava, and
  Chang}]{Alzantot2018}
Alzantot M, Sharma Y, Elgohary A, Ho BJ, Srivastava MB, Chang KW (2018)
  {Generating natural language adversarial examples}. In: Proceedings of EMNLP,
  \doi{10.18653/v1/d18-1316}

\bibitem[{Anaby-Tavor et~al.(2020)Anaby-Tavor, Carmeli, Goldbraich, Kantor,
  Kour, Shlomov, Tepper, and Zwerdling}]{Anaby-Tavor2020}
Anaby-Tavor A, Carmeli B, Goldbraich E, Kantor A, Kour G, Shlomov S, Tepper N,
  Zwerdling N (2020) {Do Not Have Enough Data? Deep Learning to the Rescue!}
  Proceedings of the AAAI \urlprefix\url{http://arxiv.org/abs/1911.03118}

\bibitem[{Banko and Brill(2001)}]{Banko2001}
Banko M, Brill E (2001) {Scaling to very very large corpora for natural
  language disambiguation}. In: Proceedings of the 39th annual meeting of the
  Association for Computational Linguistics, \doi{10.3115/1073012.1073017}

\bibitem[{Bayer et~al.(2021)Bayer, Kaufhold, and
  Reuter}]{bayerSurveyDataAugmentation2021}
Bayer M, Kaufhold MA, Reuter C (2021) A {Survey} on {Data} {Augmentation} for
  {Text} {Classification}. arXiv

\bibitem[{Belinkov and Bisk(2018)}]{Belinkov2018}
Belinkov Y, Bisk Y (2018) {Synthetic and natural noise both break neural
  machine translation}. In: Proceedings of ICLR

\bibitem[{Brown et~al.(2020)Brown, Mann, Ryder, Subbiah, Kaplan, Dhariwal,
  Neelakantan, Shyam, Sastry, Askell, Agarwal, Herbert-Voss, Krueger, Henighan,
  Child, Ramesh, Ziegler, Wu, Winter, Hesse, Chen, Sigler, Litwin, Gray, Chess,
  Clark, Berner, McCandlish, Radford, Sutskever, and Amodei}]{Brown2020}
Brown TB, Mann B, Ryder N, Subbiah M, Kaplan J, Dhariwal P, Neelakantan A,
  Shyam P, Sastry G, Askell A, Agarwal S, Herbert-Voss A, Krueger G, Henighan
  T, Child R, Ramesh A, Ziegler DM, Wu J, Winter C, Hesse C, Chen M, Sigler E,
  Litwin M, Gray S, Chess B, Clark J, Berner C, McCandlish S, Radford A,
  Sutskever I, Amodei D (2020) {Language Models are Few-Shot Learners}. In:
  NeurIPS, \urlprefix\url{http://arxiv.org/abs/2005.14165}

\bibitem[{Carreira et~al.(2004)Carreira, Crato, Gon{\c{c}}alves, and
  Jorge}]{Carreira2004}
Carreira R, Crato JM, Gon{\c{c}}alves D, Jorge JA (2004) {Evaluating adaptive
  user profiles for news classification}. In: Proceedings IUI,
  \doi{10.1145/964442.964481}

\bibitem[{Chawla et~al.(2002)Chawla, Bowyer, Hall, and Kegelmeyer}]{Chawla2002}
Chawla NV, Bowyer KW, Hall LO, Kegelmeyer WP (2002) {SMOTE: Synthetic minority
  over-sampling technique}. JAIR \doi{10.1613/jair.953}

\bibitem[{Coulombe(2018)}]{Coulombe2018}
Coulombe C (2018) {Text Data Augmentation Made Simple By Leveraging NLP Cloud
  APIs}. In: arXiv preprint arXiv:1812.04718, pp 1--33,
  \urlprefix\url{http://arxiv.org/abs/1812.04718}

\bibitem[{Fadaee et~al.(2017)Fadaee, Bisazza, and Monz}]{Fadaee2017}
Fadaee M, Bisazza A, Monz C (2017) {Data augmentation for low-Resource neural
  machine translation}. In: ACL, \doi{10.18653/v1/P17-2090}

\bibitem[{Howard and Gugger(2020)}]{Howard2020}
Howard J, Gugger S (2020) {Fastai: A layered api for deep learning}.
  Information (Switzerland) \doi{10.3390/info11020108}

\bibitem[{Howard and Ruder(2018)}]{Howard2018}
Howard J, Ruder S (2018) {Universal language model fine-tuning for text
  classification}. In: Proceedings of ACL, \doi{10.18653/v1/p18-1031}

\bibitem[{Hu and Yu(2020)}]{Hu2020}
Hu YQ, Yu Y (2020) A technical view on neural architecture search.
  International Journal of Machine Learning and Cybernetics 11(4):795--811,
  \doi{10.1007/s13042-020-01062-1}

\bibitem[{Hu et~al.(2019)Hu, Tan, Salakhutdinov, Mitchell, and Xing}]{Hu2019}
Hu Z, Tan B, Salakhutdinov R, Mitchell T, Xing EP (2019) {Learning data
  manipulation for augmentation and weighting}

\bibitem[{Huong and Hoang(2020)}]{Huong2020}
Huong TH, Hoang VT (2020) {A data augmentation technique based on text for
  Vietnamese sentiment analysis}. Proceedings of IAIT pp 1--5,
  \doi{10.1145/3406601.3406618}

\bibitem[{Imran et~al.(2018)Imran, Castillo, Diaz, and Vieweg}]{Imran2018}
Imran M, Castillo C, Diaz F, Vieweg S (2018) Processing social media messages
  in mass emergency: Survey summary. In: Companion Proceedings of the The Web
  Conference 2018, International World Wide Web Conferences Steering Committee,
  Republic and Canton of Geneva, CHE, WWW '18, p 507–511,
  \doi{10.1145/3184558.3186242}

\bibitem[{Jiao et~al.(2019)Jiao, Yin, Shang, Jiang, Chen, Li, Wang, and
  Liu}]{Jiao2019}
Jiao X, Yin Y, Shang L, Jiang X, Chen X, Li L, Wang F, Liu Q (2019) {TinyBERT:
  Distilling BERT for Natural Language Understanding}. In: EMNLP 2020, pp
  1--14, \urlprefix\url{http://arxiv.org/abs/1909.10351}

\bibitem[{Kafle et~al.(2018)Kafle, Yousefhussien, and Kanan}]{Kafle2018}
Kafle K, Yousefhussien M, Kanan C (2018) {Data Augmentation for Visual Question
  Answering}. In: Proceedings of the 10th International Conference on Natural
  Language Generation, \doi{10.18653/v1/w17-3529}

\bibitem[{Kaufhold(2021)}]{kaufhold_information_2021}
Kaufhold MA (2021) Information {Refinement} {Technologies} for {Crisis}
  {Informatics}: {User} {Expectations} and {Design} {Principles} for {Social}
  {Media} and {Mobile} {Apps}. Springer Vieweg, Wiesbaden, Germany,
  \doi{10.1007/978-3-658-33341-6},
  \urlprefix\url{https://www.springer.com/gp/book/9783658333430}

\bibitem[{Kaufhold et~al.(2020)Kaufhold, Bayer, and Reuter}]{Kaufhold2020}
Kaufhold MA, Bayer M, Reuter C (2020) {Rapid relevance classification of social
  media posts in disasters and emergencies: A system and evaluation featuring
  active, incremental and online learning}. Information Processing {\&}
  Management \doi{10.1016/j.ipm.2019.102132}

\bibitem[{Khan(2019)}]{Khan2019}
Khan B (2019) {Generate your own text with OpenAI's GPT-2}.
  \urlprefix\url{https://www.kaggle.com/bkkaggle/generate-your-own-text-with-openai-s-gpt-2-117m}

\bibitem[{Kingma and Ba(2015)}]{Kingma2015}
Kingma DP, Ba JL (2015) {Adam: A method for stochastic optimization}. In: ICLR
  2015 - Conference Track Proceedings

\bibitem[{Kobayashi(2018)}]{Kobayashi2018}
Kobayashi S (2018) {Contextual Augmentation: Data Augmentation by Words with
  Paradigmatic Relations}. In: arXiv preprint arXiv:1805.06201,
  \doi{10.18653/v1/n18-2072}

\bibitem[{Kolomiyets et~al.(2011)Kolomiyets, Bethard, and
  Moens}]{Kolomiyets2011}
Kolomiyets O, Bethard S, Moens MF (2011) {Model-portability experiments for
  textual temporal analysis}. In: Proceedings of ACL-HLT

\bibitem[{Krishnalal et~al.(2010)Krishnalal, Rengarajan, and
  Srinivasagan}]{Krishnalal2010}
Krishnalal G, Rengarajan SB, Srinivasagan KG (2010) {A New Text Mining Approach
  Based on HMM-SVM for Web News Classification}. International Journal of
  Computer Applications \doi{10.5120/395-589}

\bibitem[{Kruspe et~al.(2018)Kruspe, Kersten, Wiegmann, Stein, and
  Klan}]{Kruspe2018}
Kruspe A, Kersten J, Wiegmann M, Stein B, Klan F (2018) {Classification of
  Incident-related Tweets : Tackling Imbalanced Training Data using Hybrid CNNs
  and Translation-based Data Augmentation}. In: Notebook papers of TREC

\bibitem[{Kumar et~al.(2019)Kumar, Bhattamishra, Bhandari, and
  Talukdar}]{Kumar2019}
Kumar A, Bhattamishra S, Bhandari M, Talukdar P (2019) {Submodular
  optimization-based diverse paraphrasing and its effectiveness in data
  augmentation}. In: Proceedings of NAACL-HLT, pp 3609--3619,
  \doi{10.18653/v1/n19-1363}

\bibitem[{Kumar et~al.(2020)Kumar, Choudhary, and Cho}]{Kumar2020}
Kumar V, Choudhary A, Cho E (2020) {Data Augmentation using Pre-trained
  Transformer Models}

\bibitem[{LeCun et~al.(1998)LeCun, Bottou, Bengio, and Haffner}]{LeCun1998}
LeCun Y, Bottou L, Bengio Y, Haffner P (1998) {Gradient-based learning applied
  to document recognition}. Proceedings of the IEEE \doi{10.1109/5.726791}

\bibitem[{Liu and Zhang(2012)}]{Liu2012}
Liu B, Zhang L (2012) A Survey of Opinion Mining and Sentiment Analysis,
  Springer US, Boston, MA, pp 415--463. \doi{10.1007/978-1-4614-3223-4_13}

\bibitem[{Longpre et~al.(2020)Longpre, Wang, and DuBois}]{Longpre2020}
Longpre S, Wang Y, DuBois C (2020) {How Effective is Task-Agnostic Data
  Augmentation for Pretrained Transformers?} In: Findings of EMNLP

\bibitem[{Medhat et~al.(2014)Medhat, Hassan, and Korashy}]{MEDHAT20141093}
Medhat W, Hassan A, Korashy H (2014) Sentiment analysis algorithms and
  applications: A survey. Ain Shams Engineering Journal 5(4):1093--1113,
  \doi{https://doi.org/10.1016/j.asej.2014.04.011}

\bibitem[{Merity et~al.(2018)Merity, Keskar, and Socher}]{Merity2018}
Merity S, Keskar NS, Socher R (2018) {Regularizing and optimizing LSTM language
  models}. In: ICLR 2018 - Conference Track Proceedings

\bibitem[{Miyato et~al.(2017)Miyato, Dai, and Goodfellow}]{Miyato2017}
Miyato T, Dai AM, Goodfellow I (2017) {Adversarial training methods for
  semi-supervised text classification}. In: Conference Track - ICLR

\bibitem[{Nguyen et~al.(2017)Nguyen, Ali Al~Mannai, Joty, Sajjad, Imran, and
  Mitra}]{Nguyen_2017}
Nguyen D, Ali Al~Mannai K, Joty S, Sajjad H, Imran M, Mitra P (2017) Robust
  classification of crisis-related data on social networks using convolutional
  neural networks. Proceedings of the International AAAI Conference on Web and
  Social Media 11(1),
  \urlprefix\url{https://ojs.aaai.org/index.php/ICWSM/article/view/14950}

\bibitem[{Olteanu et~al.(2015)Olteanu, Vieweg, and Castillo}]{Olteanu2015}
Olteanu A, Vieweg S, Castillo C (2015) {What to expect when the unexpected
  happens: Social media communications across crises}. In: Proceedings of CSCW,
  \doi{10.1145/2675133.2675242}

\bibitem[{Qiu et~al.(2020)Qiu, Xu, Zhang, Wang, Shen, de~Melo, Long, and
  Li}]{Qiu2020}
Qiu S, Xu B, Zhang J, Wang Y, Shen X, de~Melo G, Long C, Li X (2020) {EasyAug:
  An Automatic Textual Data Augmentation Platform for Classification Tasks}.
  In: Companion Proceedings of the Web Conference 2020,
  \doi{10.1145/3366424.3383552}

\bibitem[{Radford et~al.(2018)Radford, Wu, Child, Luan, Amodei, and
  Sutskever}]{Radford2018}
Radford A, Wu J, Child R, Luan D, Amodei D, Sutskever I (2018) {Language Models
  are Unsupervised Multitask Learners}. In: OpenAI blog

\bibitem[{Raghuwanshi and Shukla(2021)}]{Raghuwanshi2021}
Raghuwanshi BS, Shukla S (2021) Classifying imbalanced data using {SMOTE} based
  class-specific kernelized {ELM}. International Journal of Machine Learning
  and Cybernetics 12(5):1255--1280, \doi{10.1007/s13042-020-01232-1}

\bibitem[{Reimers and Gurevych(2019)}]{Reimers2019}
Reimers N, Gurevych I (2019) {Sentence-BERT: Sentence Embeddings using Siamese
  BERT-Networks}. In: Proceedings of the 2019 Conference on Empirical Methods
  in Natural Language Processing and the 9th International Joint Conference on
  Natural Language Processing (EMNLP-IJCNLP), \doi{10.18653/v1/d19-1410}

\bibitem[{Reuter et~al.(2012)Reuter, Marx, and Pipek}]{reuter_crisis_2012}
Reuter C, Marx A, Pipek V (2012) Crisis {Management} 2.0: {Towards} a
  {Systematization} of {Social} {Software} {Use} in {Crisis} {Situations}.
  International Journal of Information Systems for Crisis Response and
  Management (IJISCRAM) 4(1):1--16, \doi{10.4018/jiscrm.2012010101}

\bibitem[{Reuter et~al.(2016)Reuter, Ludwig, Kaufhold, and
  Spielhofer}]{reuter_emergency_2016}
Reuter C, Ludwig T, Kaufhold MA, Spielhofer T (2016) Emergency {Services}
  {Attitudes} towards {Social} {Media}: {A} {Quantitative} and {Qualitative}
  {Survey} across {Europe}. International Journal on Human-Computer Studies
  (IJHCS) 95:96--111, \doi{10.1016/j.ijhcs.2016.03.005}

\bibitem[{Rizos et~al.(2019)Rizos, Hemker, and Schuller}]{Rizos2019}
Rizos G, Hemker K, Schuller B (2019) {Augment to prevent: Short-text data
  augmentation in deep learning for hate-speech classification}. In:
  Proceedings of CIKM, \doi{10.1145/3357384.3358040}

\bibitem[{Schulz et~al.(2017)Schulz, Guckelsberger, and Janssen}]{Schulz2017}
Schulz A, Guckelsberger C, Janssen F (2017) {Semantic Abstraction for
  generalization of tweet classification: An evaluation of incident-related
  tweets}. Semantic Web \doi{10.3233/SW-150188}

\bibitem[{Sennrich et~al.(2016)Sennrich, Haddow, and Birch}]{Sennrich2016}
Sennrich R, Haddow B, Birch A (2016) {Improving neural machine translation
  models with monolingual data}. In: ACL, \doi{10.18653/v1/p16-1009}

\bibitem[{Shorten and Khoshgoftaar(2019)}]{Shorten2019}
Shorten C, Khoshgoftaar TM (2019) {A survey on Image Data Augmentation for Deep
  Learning}. Journal of Big Data \doi{10.1186/s40537-019-0197-0}

\bibitem[{Smith(2018)}]{Smith2018}
Smith LN (2018) {A disciplined approach to neural network hyper-parameters:
  Part 1 -- learning rate, batch size, momentum, and weight decay}

\bibitem[{Socher et~al.(2013)Socher, Perelygin, Wu, Chuang, Manning, Ng, and
  Potts}]{Socher2013}
Socher R, Perelygin A, Wu JY, Chuang J, Manning CD, Ng AY, Potts C (2013)
  {Recursive deep models for semantic compositionality over a sentiment
  treebank}. In: Proceedings of EMNLP

\bibitem[{Soden and Palen(2018)}]{Soden2018}
Soden R, Palen L (2018) Informating crisis: Expanding critical perspectives in
  crisis informatics. Proc ACM Hum-Comput Interact 2(CSCW),
  \doi{10.1145/3274431}

\bibitem[{Solaiman et~al.(2019)Solaiman, Brundage, Clark, Askell, Herbert-Voss,
  Wu, Radford, and Wang}]{Solaiman2019}
Solaiman I, Brundage M, Clark J, Askell A, Herbert-Voss A, Wu J, Radford A,
  Wang J (2019) {Release strategies and the social impacts of language models}

\bibitem[{Stieglitz et~al.(2018)Stieglitz, Mirbabaie, Ross, and
  Neuberger}]{stieglitzSocialMediaAnalytics2018}
Stieglitz S, Mirbabaie M, Ross B, Neuberger C (2018) Social media analytics –
  {{Challenges}} in topic discovery, data collection, and data preparation.
  International Journal of Information Management 39:156--168

\bibitem[{Sun et~al.(2017)Sun, Shrivastava, Singh, and Gupta}]{Sun2017}
Sun C, Shrivastava A, Singh S, Gupta A (2017) {Revisiting Unreasonable
  Effectiveness of Data in Deep Learning Era}. In: Proceedings of the ICCV,
  \doi{10.1109/ICCV.2017.97}

\bibitem[{Sun and He(2020)}]{Sun2020}
Sun X, He J (2020) {A novel approach to generate a large scale of supervised
  data for short text sentiment analysis}. Multimedia Tools and Applications
  \doi{10.1007/s11042-018-5748-4}

\bibitem[{Taylor and Nitschke(2019)}]{Taylor2019}
Taylor L, Nitschke G (2019) {Improving Deep Learning with Generic Data
  Augmentation}. In: Proceedings of SSCI, \doi{10.1109/SSCI.2018.8628742}

\bibitem[{Wang and Lillis(2020)}]{Wang2020}
Wang C, Lillis D (2020) {Classification for Crisis-Related Tweets Leveraging
  Word Embeddings and Data Augmentation}. In: {TREC 2019},
  \urlprefix\url{https://trec.nist.gov/}

\bibitem[{Wang and Yang(2015)}]{Wang2015}
Wang WY, Yang D (2015) {That's so annoying!!!: A lexical and frame-semantic
  embedding based data augmentation approach to automatic categorization of
  annoying behaviors using {\#}petpeeve tweets}. In: Proceedings of EMNLP,
  \doi{10.18653/v1/d15-1306}

\bibitem[{Wei and Zou(2019)}]{Wei2019}
Wei J, Zou K (2019) {EDA: Easy Data Augmentation Techniques for Boosting
  Performance on Text Classification Tasks}. In: Proceedings of the 2019
  Conference on Empirical Methods in Natural Language Processing and the 9th
  International Joint Conference on Natural Language Processing (EMNLP-IJCNLP),
  \doi{10.18653/v1/d19-1670}

\bibitem[{Woolf(2019)}]{Woolf2019}
Woolf M (2019) {GitHub - gpt-2-simple: Python package to easily retrain
  OpenAI's GPT-2 text-generating model on new texts}.
  \urlprefix\url{https://github.com/minimaxir/gpt-2-simple}

\bibitem[{Xiang et~al.(2021)Xiang, Chersoni, Lu, Huang, Li, and
  Long}]{xiangLexicalDataAugmentation2021}
Xiang R, Chersoni E, Lu Q, Huang CR, Li W, Long Y (2021) Lexical data
  augmentation for sentiment analysis. Journal of the Association for
  Information Science and Technology 72(11):1432--1447,
  \doi{10.1002/asi.24493},
  \urlprefix\url{https://onlinelibrary.wiley.com/doi/abs/10.1002/asi.24493},
  \_eprint: https://onlinelibrary.wiley.com/doi/pdf/10.1002/asi.24493

\bibitem[{Xu et~al.(2016)Xu, Jia, Mou, Li, Chen, Lu, and Jin}]{Xu2016}
Xu Y, Jia R, Mou L, Li G, Chen Y, Lu Y, Jin Z (2016) {Improved relation
  classification by deep recurrent neural networks with data augmentation}. In:
  Proceedings of COLING 2016: Technical Papers

\bibitem[{Zeiler and Fergus(2013)}]{Zeiler2013}
Zeiler MD, Fergus R (2013) {Stochastic pooling for regularization of deep
  convolutional neural networks}. In: Proceedings of ICLR

\bibitem[{Zhai et~al.(2021)Zhai, Qi, and Zhang}]{Zhai_2021}
Zhai J, Qi J, Zhang S (2021) Imbalanced data classification based on diverse
  sample generation and classifier fusion. International Journal of Machine
  Learning and Cybernetics \doi{10.1007/s13042-021-01321-9}

\bibitem[{Zhang et~al.(2018)Zhang, Cisse, Dauphin, and Lopez-Paz}]{Zhang2018}
Zhang H, Cisse M, Dauphin YN, Lopez-Paz D (2018) {MixUp: Beyond empirical risk
  minimization}. In: Conference Track of ICLR

\bibitem[{Zhang et~al.(2015)Zhang, Zhao, and Lecun}]{Zhang2015a}
Zhang X, Zhao J, Lecun Y (2015) {Character-level convolutional networks for
  text classification}. In: NIPS

\end{thebibliography}

%
%

\section*{Appendices} \label{appendix}
\addcontentsline{toc}{section}{Appendices}
\renewcommand{\thesubsection}{\Alph{subsection}}
\setcounter{subsection}{0}

\subsection{Availability of Data and Material} \label{appendix_data_material}
In summary, most datasets analyzed during the current study are publicly available. In the \textbf{first evaluation} (section \ref{results_1}), we used the SST dataset\footnote{SST datasets of \citet{Socher2013}: \url{https://nlp.stanford.edu/sentiment/index.html}} of \citet{Socher2013} which is publicly available.

The datasets analyzed during the \textbf{second evaluation} (section \ref{results_2}) are not publicly available due to publication restrictions by news agencies. Still, a concise description of these datasets is given in appendix \ref{appendix_dataset}. We would like to highlight that news classification is a highly relevant field in the industry that receives too little attention in academia. In addition, we had to decide to create our own dataset in order to tackle long text classification, as these datasets are particularly rare.  We hope that we have included enough results from the public datasets so that reproducibility can be inferred.

For the \textbf{third evaluation} (section \ref{results_3}), three public datasets (Boston Bombings, the Bohol Earthquake and the West Texas Explosions) from the CrisisLexT26 \footnote{CrisisLexT26 datasets from \citet{Olteanu2015}: \url{https://github.com/sajao/CrisisLex/tree/master/data/CrisisLexT26}} annotated data groups from \citet{Olteanu2015} and two public datasets (Dublin and New York City) from the annotated data groups from \citet{Schulz2017}\footnote{Datasets from \citet{Schulz2017}: \url{ http://www.doc.gold.ac.uk/~cguck001/IncidentTweets/}}  were used. The primary language of all datasets is English.


\subsection{Description of the Datasets Used During the Second Evaluation} \label{appendix_dataset}
The datasets consist of English news articles of over 2,600 different source domains from the years 2019 and 2020, which were preselected with regard to specific query words. For each topic, the articles received values on the basis of these query words to split them into 12 different buckets. We sampled the news articles from these buckets uniformly so that highly diverse instances were labeled. A short summary of the labeling guidelines and data distributions are described in the following list:

\textbf{Layoff.} The layoff topic consists of all forms of dismissals of employees in the corporate context. A total of 1992 articles were annotated of which 751 instances are positive and 1241 are negative.

\textbf{Management change.} This topic covers all forms of changes (retirement, resignation, appointment) of the board of directors and important positions in companies, organizations and advisory boards. 2129 instances were labeled of which 567 instances are positive and 1562 are negative.

\textbf{Mergers and Acquisitions.} M\&A includes all transactions in which ownership is transferred to companies or their operating units. The mere investment in a company is seen as negative here. For this topic, 2227 instances were labeled of which 474 are positive and 1753 are negative.

\textbf{Flood.} The flooding topic is positively recognized in news if the article deals with the actual flooding or the main topic is a consequence of a flood. If only an increase in the water level is reported, this message should not be regarded as positive. There are 2533 identified instances on this topic, 1639 of which are positive and 894 are negative. Although more positive than negative instances were identified here, the negative class is still seen as the majority class since the clear majority of all messages on the Internet are not related to floods. 

\textbf{Wildfire.} All forms of wildfire are classified under this topic. House fires and metaphorical uses of the term are labeled negatively. The dataset includes 2410 identified instances of which 1202 are positive and 1208 are negative. Similar to the flood topic, the negative class is the majority class.

In the preprocessing step we added the tokens “xxtitle” before every title and “xxbodytext” before the start of the normal article text.

\subsection{Algorithm} \label{appendix_algorithm}
In the following, we defined the algorithm of the context independent data augmentation method.

\begin{algorithm}
\SetAlgoLined
\SetKwInOut{Input}{Input}
\Input{Language Model $LM$, Class data $X_c$,\\ Document Embedding Model $E$,\\ Number of instances per training data $n$}
1. For each instance in $X_c$: Attach `$<|startoftext|> |i|$' to the beginning and `$<|endoftext|>$' to the end of the $i$th instance to obtain $X_{cprep}$\\
2. Finetune $LM$ on $X_{cprep}$ to obtain $LM_{cprep}$\\
3. For each k in $|X_{cprep}|$: Generate $n$ new instances with $LM_{cprep}$  and `$<|startoftext|> |i|$' as prefix to obtain $X_{gen}$\\
4. Embed all instances in $X_{gen}$ and $X_{c}$ with $E$ \\
5. Obtain $X_{filtered}$ by including all instances of $X_{gen}$ at which the embedding representations are close to the centroid of the embedded $X_{c}$

\KwResult{$X_{filtered}$ }
 \caption{Augmentation of Short Texts}
\end{algorithm}

\begin{algorithm}
\SetAlgoLined
\SetKwInOut{Input}{Input}
\Input{Language Model $LM$, Class data $X_c$,\\ Document Embedding Model $E$,\\ Number of instances per training data $n$, \\ Function extracting context part $cont()$}
1. For each instance in $X_c$: Attach `$<|startoftext|>$' to the beginning and `$<|endoftext|>$' to the end of the instances to obtain $X_{cprep}$\\
2. Finetune $LM$ on $X_{cprep}$ to obtain $LM_{cprep}$\\
3. For each $t$ in $X_{cprep}$: Generate $n$ new instances with $LM_{cprep}$  and `$<|startoftext|>$' + $cont(t)$ as prefix to obtain $X_{gen}$\\
4. Embed all instances in $X_{gen}$ and $X_{c}$ with $E$ \\
5. Obtain $X_{filtered}$ by including all instances of $X_{gen}$ at which the embedding representations are close to the centroid of the embedded $X_{c}$

\KwResult{$X_{filtered}$ }
 \caption{Augmentation of Long Texts}
 
\end{algorithm}
\subsection{Ethics} \label{appendix_ethics}
In our work, we have placed particular emphasis on ethics and repeatedly reassessed our approach with regard to responsibilities. We have restricted ourselves to only using the textual content and the labels in the datasets to respect privacy as much as possible. When working with social media data, we especially did not use, aggregate or draw any conclusion from further metadata such as the name or location of a user. 

For the practical implementation of our method, we want to mention that the GPT-2 model as well as most other language models contain biases (for example a gender, religious or racial bias) \citep{Solaiman2019}. Using the method in a real application can result in a domain shift and/or an inclusion of those biases. This can explicitly lead to discriminatory decisions by the machine learning model, even if the dataset itself does not contain any bias.

\subsection{Architecture, Hyperparameters and Infrastructure} \label{appendix_architecture}
In the evaluation, we used the pre-trained ULMFit model by \citet{Howard2018} as implemented in fastai \citep{Howard2020}. The model consists of a pre-trained encoder that is based on the AWD-LSTM architecture by \citet{Merity2018} and a linear pooling classifier. The classifier consists of a layer that concatenates the final outputs of the encoder with the maximum and the average of all intermediate outputs and two fully connected layers\footnote{ULMFit implementation: \url{https://fastai1.fast.ai/text.learner.html\#text_classifier_learner}}. Further information about the general architecture can be extracted from the paper of \citet{Howard2018} and the implementation in fastai \citep{Howard2020}. 
The encoder finetuning is done by preparing all available data of the respective task (including the augmented data) for the language modeling task. In the training, we performed 15 cycles with the 1cycle policy by \citet{Smith2018}. We used a learning rate of 0.002 and a maximum and minimum momentum of 0.8 and 0.7. Overall, we used a fixed batch size of 64 and a backpropagation through time window of 70. Each encoder and classifier was trained to the downstream task with three cycles with gradual unfreezing and another five cycles with the unfrozen model. The learning rate was individually determined by the learning rate range test by \citet{Smith2018} with a range from $10^{-7}$ to 10 over 100 iterations\footnote{Learning rate range test implementation: \url{https://fastai1.fast.ai/callbacks.lr_finder.html\#callbacks.lr_finder}}. On the outputs we used label smoothing with an epsilon parameter of 0.1. As optimization algorithm, we used the Adam algorithm \citep{Kingma2015}.

For the text generation, we used the gpt-2-simple implementation\footnote{gpt-2-simple implementation:
\url{https://github.com/minimaxir/gpt-2-simple}}  to finetune and generate texts from the GPT-2 model. Only the parameters discussed in chapter 4.2 were varied. For the filtering step, Sentence-BERT\footnote{Sentence-BERT implementation: \url{https://github.com/UKPLab/sentence-transformers}}  \citep{Reimers2019} with the “roberta-large-nli-stsb-mean-tokens” transformer model was used.

We tokenized all datasets with the built-in tokenizers of the gpt-2-simple, Sentence-BERT and fastai libraries for the different use cases.

The evaluation of chapter 4.4 was performed on a Nvidia Quadro RTX 6000 graphics card with 24 GB RAM. For the evaluation of chapter 4.3 and 4.5, less resources were necessary, which is why a Nvidia Tesla P100 with 16GB RAM was used.

In general, the tasks were quite resource intensive. The finetuning of the GPT-2 model on the flood dataset took about five hours, while the generation of ten examples per instance occupied about four days. The finetuning of the language model and the training of the classifier together took about another four hours.

\subsection{Generated Data} \label{appendix_data}

\begin{table*}[t]
\centering
\caption{Generated data instances and their most similar original counterparts.}
\label{table:generated_data}
\begin{tabularx}{\textwidth}{lX}
\hline
\multirow{10}{*}{\begin{tabular}[c]{@{}c@{}}West Texas \\ Explosions\end{tabular}} &
  Media Matters Texas:  West, Texas, fertilizer plant: 28 years, no full inspection: Before 270 tons of ammonium nitratet exploded at ... http://t.co/XwUwgRcGDT \\
 &
  West, Texas, fertilizer plant: 28 years, no full inspection: Before 270 tons of ammonium nitratet exploded at ... http://t.co/XwUwgRcGDT \\ \cline{2-2} 
 &
  RT @mpoindc: I want them to know that they are not forgotten, Obama says to victims of explosion in West, TX. \\
 &
  I want them to know that they are not forgotten, Obama says to victims of explosion in West, TX. \\ \cline{2-2} 
 &
  RT @BreakingNews: Large explosion reported at fertilizer plant near Waco, Texas - @CBSDFW http://t.co/xOIyCyuxFD \\
 &
  Explosion reported at fertilizer plant near Waco, Texas - @CBSDFW http://t.co/xOIyCyuxFD \\ \cline{2-2} 
 &
  RT @BBCBreaking: 12 confirmed dead, approximately 200 injured in \#West fertiliser plant explosion in Texas, say state officials \\
 &
  Searchers Find 12 Bodies After Texas Explosion http://t.co/fnBc2LmXs (CNN) — Hundreds believed injured in Texas fertilizer plant explosion, medical examiner says. http://t ... \\ \cline{2-2} 
 &
  RT @laurenonizzle: This iPhone video of the fertilizer plant explosion near \#Waco will send chills up your spine. Surreal. http://t.co/O ... \\
 &
  RT @laurenonizzle: This iPhone video of the fertilizer plant explosion near \#Waco will send chills up your spine. Surreal. http://t.co/O ... \\ \cline{1-2} 
\multirow{10}{*}{SST-2 (100)} &
  smart , sassy interpretation of the oscar wilde play . \\
 &
  harp , sassy interpretation of the oscar wilde play , with an unexpected twist . \\ \cline{2-2} 
 &
  a fast paced and suspenseful argentinian thriller about the shadow side of play . \\
 &
  a fast paced and suspenseful argentinian thriller . \\ \cline{2-2} 
 &
  this comic gem is as delightful as it is derivative . \\
 &
  the film is bright and flashy in all the right ways . \\ \cline{2-2} 
 &
  the best movie in many a moon about the passions that sometimes fuel our best achievements and other times leave us stranded with nothing more than our lesser appetites . \\
 &
  the best film in many a moon about the passions that sometimes fuel our best achievements and other times leave us stranded with nothing more than our lesser appetites . \\ \cline{2-2} 
 &
  a fine production with splendid singing by angela gheorghiu , ruggero raimondi , and roberto alagna . \\
 &
  a fine production with splendid singing by angela gheorghiu , ruggero raimondi , and roberto alagna . \\ \cline{1-2} 
\end{tabularx}
\end{table*}
In our experiments, we also analyzed the generated data. For this purpose, we first selected generated instances and tried to find the original instance with the closest resemblance (measured by Levenshtein distance). In Table \ref{table:generated_data} a excerpt of some instances, original and generated counterparts, from the West Texas Explosions and SST-2 datasets are given. Here we can see, that the GPT-2 model is, for example, able to remove preceding words (first and second instance of West Texas Explosions) or even truncate the sentence at the end (second instance of SST-2). The first example of the SST-2 dataset shows that the model is also able to enlarge the original instance. In case of the third example of West Texas Explosions it could be that the original instance is interpolated with another instance. The fourth instance of the SST-2 dataset also shows that the model can make small changes like synonym substitution. Furthermore, for many generated instances we were not able to find similar counterparts, see for example the third instance of the SST-2 dataset given in Table \ref{table:generated_data}. These can, for example, be instances the model learned beforehand, which supports the consideration that the model is able to create highly diverse examples with new linguistic features. However, as shown with the last examples of the two datasets, the model also sometimes repeats the original instance. We also noticed that the more data we generate, the more duplicates can be found. This is reasonable as the probability for some token sequences is very high and they are therefore generated more often. As already indicated to some extent in the ablation studies, it would be interesting to see how many artificial instances can be generated till a maximum improvement is reached and whether this is dependent on the size of the language model. In general, the findings indicate that the method proposed in our study is capable of performing many different transformations.

\end{document}